%% file: vbmnn.tex
\documentclass{article}

\usepackage{times}
\usepackage{graphicx} 
\usepackage{subfigure} 
\usepackage{dblfloatfix}
\usepackage{natbib}

\usepackage{algorithm}
\usepackage{algorithmic}

\usepackage{hyperref}

\usepackage[accepted]{icml2016}

\icmltitlerunning{Structured and Efficient Variational Deep Learning with Matrix Gaussian Posteriors}

\usepackage{amssymb} 
\usepackage{amsmath} 
\DeclareMathOperator{\E}{\mathbb{E}}

\DeclareMathOperator{\vect}{vec}
\DeclareMathOperator{\tr}{tr}
\def\!#1{\boldsymbol{#1}}
\def\*#1{\mathbf{#1}}

\begin{document} 

\twocolumn[
\icmltitle{Structured and Efficient Variational Deep Learning with Matrix Gaussian Posteriors}

\icmlauthor{Christos Louizos}{c.louizos@uva.nl}
\icmladdress{AMLAB, Informatics Institute, University of Amsterdam}
\icmlauthor{Max Welling}{m.welling@uva.nl}
\icmladdress{AMLAB, Informatics Institute, University of Amsterdam \\
Canadian Institute for Advanced Research (CIFAR)}

\icmlkeywords{Bayesian Neural Networks, Deep Learning, Gaussian Process}

\vskip 0.3in
]

\begin{abstract} 
We introduce a variational Bayesian neural network where the parameters are governed via a probability distribution on random matrices. Specifically, we employ a matrix variate Gaussian~\cite{gupta1999matrix} parameter posterior distribution where we explicitly model the covariance among the input and output dimensions of each layer. Furthermore, with approximate covariance matrices we can achieve a more efficient way to represent those correlations that is also cheaper than fully factorized parameter posteriors. We further show that with the ``local reprarametrization trick"~\cite{kingma2015variational} on this posterior distribution we arrive at a Gaussian Process~\cite{rasmussen2006gaussian} interpretation of the hidden units in each layer and we, similarly with~\cite{gal2015dropout}, provide connections with deep Gaussian processes. We continue in taking advantage of this duality and incorporate ``pseudo-data''~\cite{snelson2005sparse} in our model, which in turn allows for more efficient posterior sampling while maintaining the properties of the original model. The validity of the proposed approach is verified through extensive experiments.
\end{abstract} 

\section{Introduction}
\label{sec:intro}
While deep learning methods are beating every record in terms of predictive accuracy, they do not yet provide the user with reliable confidence intervals. Yet, for most applications where \emph{decisions} are made based on these predictions, confidence intervals are key. Take the example of an autonomous driving vehicle that enters a new unknown traffic situation: recognizing that predictions become unreliable and handing the steering wheel back to the driver is essential. Similarly, when a physician diagnoses a patient with some ailment and prescribes a drug with potentially severe side effects, it is essential that she/he knows when predictions are unreliable and additional investigation is necessary. These considerations have motivated us to develop a fully Bayesian deep learning framework that is accurate, efficient and delivers reliable confidence intervals.  

Furthermore, by being Bayesian we can also harvest another property as a byproduct; natural protection against \emph{overfitting}. Instead of making point estimates for the parameters of the network, which can overfit and provide erroneously certain predictions, we estimate a full posterior distribution over these parameters. Armed with these posterior distributions we can now perform predictions using the posterior predictive distribution, i.e. we can now marginalize over the network parameters and make predictions on the basis of the datapoints alone. As a result we can both obtain the aforementioned confidence intervals and better regularize our networks, which is very important in problems where we do not have enough data relative to the amount of features. 

Obtaining the parameter posterior distributions for large neural networks is however intractable. To this end, many methods for approximate posterior inference have been devised. Markov Chain Monte Carlo (MCMC) methods are one class of methods that have been explored in this context via Hamiltonian Monte Carlo~\cite{neal2012bayesian} and stochastic gradient methods~\cite{welling2011bayesian,ahn2012bayesian}. 

Another family of methods that provide deterministic approximations to the posterior are based on variational inference. These cast inference as an optimization problem and minimize the KL-divergence between the approximate and true posterior. There have been many recent attempts that have adopted this paradigm~\citep{graves2011practical,hernandez2015probabilistic,blundell2015weight,kingma2015variational}. However, most of these approaches assume a fully factorized posterior distribution over the neural network weights. We conjecture that this assumption is very restricting as the ``true" posterior distribution does have some correlations among the network weights. Therefore by using a fully factorized posterior distribution the learning task becomes ``harder" as there is not enough information sharing among the weights. 

We therefore introduce a variational Bayesian neural network that instead of treating each element of the weight matrix independently, it treats the weight matrix \emph{as a whole} via a matrix variate Gaussian distribution~\cite{gupta1999matrix}, i.e. a distribution over random matrices. This parametrization will significantly reduce the amount of variance-related parameters that we have to estimate: instead of estimating a separate variance for each weight we can now estimate separate variances for each row and column of the weight matrix, i.e input and output feature specific variances. This will immediately introduce correlations, and consequently information sharing, among the weights. As a result, it will allow for an easier estimation of the weight posterior uncertainty. 

In addition, we will also provide a distinct relation between our model and deep (multi-output) Gaussian Processes~\cite{damianou2012deep}; this relation arises through the application of the ``local reparametrization trick"~\cite{kingma2015variational} on the matrix variate Gaussian distribution. This fact reveals an interesting property for this Bayesian neural network: we can now sample more efficiently while maintaining the properties of the original model through the introduction of pseudo-data~\cite{snelson2005sparse}.

\section{Beyond fully factorized parameter posteriors}
\input{model}

\section{Related work}
\input{related}

\section{Experiments}
\input{experiments}

\section{Conclusions}
\input{conclusion}

\section*{Acknowledgements} 
We would like to thank anonymous reviewers for their feedback. This research is supported by TNO, Scyfer B.V., NWO, Google and Facebook.

\bibliography{vmnn}
\bibliographystyle{icml2016}

\newpage
\clearpage
\appendix
\input{appendix}

\end{document}

%% file: model.tex
\label{sec:background}
\subsection{Matrix variate Gaussian distribution}
The matrix variate Gaussian~\cite{gupta1999matrix} is a three parameter distribution that governs a random matrix, e.g. $\*W$:
\begin{align}
    p(\*W) & = \mathcal{MN}(\*M, \*U, \*V)\nonumber\\
     & = \frac{\exp\big(-\frac{1}{2}\tr\big[\*V^{-1}(\*W - \*M)^T\*U^{-1}(\*W - \*M)\big]\big)}{(2\pi)^{np/2} |\*V|^{n/2} |\*U| ^{n/2}}
\end{align}
where $\*M$ is a $r \times c$ matrix that is the mean of the distribution, $\*U$ is a $r\times r$ matrix that provides the covariance of the rows and $\*V$ is a $c\times c$ matrix that governs the covariance of the columns of the matrix. According to~\cite{gupta1999matrix} this distribution is essentially a multivariate Gaussian distribution where:
\begin{align}
    p(\vect(\*W)) = \mathcal{N}(\vect(\*M), \*V \otimes \*U) \nonumber
\end{align}
where $\vect(\cdot)$ is the vectorization operator (i.e. stacking the columns into a single vector) and $\otimes$ is the Kronecker product. Despite the fact that the matrix variate Gaussian is a simple generalization of the multivariate case it provides us a straightforward way to separate the correlations among the rows and columns of the matrix, which implicitly affects the correlations among the input and output hidden units. 

\subsection{Variational inference  with matrix variate Gaussian posteriors}
For the following we will assume that each input to a layer is augmented with an extra dimension containing $\*1$'s so as to account for the biases and thus we are only dealing with weights $\*W$ on this expanded input. In order to obtain a matrix variate Gaussian posterior distribution for these weights we can work in a pretty straightforward way: the derivation is similar to~\citep{graves2011practical,kingma2013auto,blundell2015weight,kingma2015variational}. Let $p_\theta(\*W), q_\phi(\*W)$ be a matrix variate Gaussian prior and posterior distribution with parameters $\theta, \phi$ respectively and $(\*x_i, \*y_i)_{i=1}^{N}$ be the training data sampled from the empirical distribution $\tilde{p}(\*x, \*y)$. Then the following lower bound on the marginal log-likelihood can be derived:
\begin{alignat}{3}
    &\mathcal{L}(\phi;\theta) = \E_{\tilde{p}(\*x, \*y)}[\log p(\*Y|\*X)] \leq \nonumber\\&\quad\E_{\tilde{p}(\*x, \*y)}\bigg[\int q_\phi(\*W)\log \frac{p_\theta(\*W)p(\*Y|\*X, \*W)}{q_\phi(\*W)}d\*W\bigg]\nonumber\\
    & = \E_{\tilde{p}(\*x, \*y)}\bigg[\E_{q_\phi(\*W)}\big[\log p(\*Y|\*X, \*W)\big] - \nonumber\\&\qquad-KL(q_\phi(\*W) || p_\theta(\*W))\bigg]
    \label{eq:marginal_lb}
\end{alignat}
Following~\cite{graves2011practical,blundell2015weight,kingma2015variational} we will refer to $L_{(\*X, \*Y)} = \E_{q_\phi(\*W)}\big[\log p(\*Y|\*X,\*W)\big]$ as the \emph{expected log-likelihood} and to $L_c = -KL(q_\phi(\*W) || p_\theta(\*W))$ as the \emph{complexity loss}. To estimate $L_{(\*X, \*Y)}$ we will use simple Monte Carlo integration along with the ``reparametrization trick"~\citep{kingma2013auto, rezende2014stochastic}:
\begin{alignat}{3}
   &\E_{q_\phi(\*W)}\big[\log p(\*Y|\*X, \*W)\big] = \frac{1}{L}\sum_{i=1}^{L} \log p(\*Y| \*X, \*W^{(l)}) \label{eq:original_rep}\\
   &\*W^{(l)} = \*M + \*U^\frac{1}{2}\*E^{(l)}\*V^\frac{1}{2}\nonumber\\
   &\*E^{(l)} \sim \mathcal{MN}(\*0, \*I, \*I) \quad \text{\big(i.e. }E_{ij} \sim \mathcal{N}(0, 1)\text{\big)} \nonumber
\end{alignat}
As for the complexity loss $L_c$;  due to the relation with the multivariate Gaussian we can still calculate the KL-divergence between the matrix variate Gaussian prior and posterior efficiently in closed form. 

However, maintaining a full covariance over the rows and columns of the weight matrix is both memory and computationally intensive. In order to still have a tractable model we approximate each of the covariances with a diagonal matrix (i.e. independent rows and columns) for simplicity\footnote{Note that we could also easily use rank-1 matrices with diagonal corrections~\citep{rezende2014stochastic} and increase the flexibility of our posterior. For example we could apply the rank-1 approximation to the square root of the covariance matrix (as we directly use it for sampling), i.e. $\*C^{\frac{1}{2}} = \*D_c + \*u\*u^T$ where $\*D_c$ is a diagonal matrix with positive elements.}. This approximation provides a per-layer parametrization that requires significantly less parameters than a simple fully factorized Gaussian posterior: we have a total of $(n_{in}\times n_{out}) + n_{in} + n_{out}$ parameters, whereas a fully factorized Gaussian posterior has $2(n_{in}\times n_{out})$ per layer. This in turn makes the posterior uncertainty estimation easier as there are both fewer parameters to learn and also ``information sharing" among the weights due to the induced correlations.

With this diagonal approximation to the covariance matrices the KL-divergence between the matrix variate Gaussian posterior $q(\*W|\*M, \!\sigma_r^2\*I, \!\sigma_c^2\*I)$ and a standard isotropic matrix variate Gaussian prior $p(\*W|\*0,\*I,\*I)$ for a matrix of size $r\times c$ corresponds to the following simple expression:
\begin{alignat}{3}
&KL(q(\*W|\*M, \!\sigma_r^2\*I, \!\sigma_c^2\*I) || p(\*W|\*0,\*I,\*I)) = \nonumber \\ &\qquad\frac{1}{2}\Bigg(\bigg(\sum_{i=1}^{r}\sigma_{r_i}^2\bigg)\bigg(\sum_{j=1}^{c}\sigma_{c_j}^2\bigg) + \|\*M\|_F^2 - rc \nonumber\\&\qquad -  c \bigg(\sum_{i=1}^{r}\log\sigma_{r_i}^2\bigg) - r \bigg(\sum_{j=1}^{c}\log\sigma_{c_j}^2\bigg)\Bigg)
\end{alignat}
The derivation for arbitrary covariance matrices is given in the appendix. 

\subsection{Deep matrix variate Bayesian nets as deep multi-output Gaussian Processes}
\label{sec:local_rep}
Directly using the expected log-likelihood estimator~\ref{eq:original_rep} yields increased variance and higher memory requirements, as it was pointed in~\cite{kingma2015variational}. Fortunately, similarly to a standard multivariate Gaussian, the inner product between a matrix and a matrix variate Gaussian is again a matrix variate Gaussian~\cite{gupta1999matrix} and as a result we can use the ``local reparametrization trick''~\cite{kingma2015variational}. Let $\*A_{M\times r}$, with $M \leq r$, be a minibatch of $M$ inputs with dimension $r$ that is the input to a network layer; the inner product $\*B_{M\times c} = \*A\*W$, where $\*W$ is a matrix variate variable with size $r \times c$, has the following distribution:
\begin{align}
    p(\*B|\*A) = \mathcal{MN}(\*A\*M, \*A\*U\*A^T, \*V)
    \label{eq:mnh}
\end{align}
As we can see, after the inner product the inputs $\*A$ become \emph{dependent} due to the non-diagonal row covariance $\*A\*U\*A^T$. Furthermore, the resulting matrix variate Gaussian maintains the same marginalization properties as a multivariate Gaussian. More specifically, if we marginalize out a row from the $\*B$ matrix, then the resulting distribution depends only on the remaining inputs, i.e. it corresponds to simply removing that particular input from the minibatch. This fact exposes a Gaussian Process~\cite{rasmussen2006gaussian} nature for the output $\*B$ of each layer.

To make the connection even clearer we can consider an example similar to the one presented in~\cite{gal2015dropout}. Let's assume that we have a neural network with one hidden layer and one output layer. Furthermore, let $\*X$, with dimensions $N\times D_x$, be the input to the network and $\*Y$, with dimensions $N\times D_y$, be the target variable. Finally, let's also assume that for the first weight matrix $p_{\theta_1}(\*W_1) = \mathcal{MN}(\*0, \*U^0_1, \*V^0_1)$ and that for the second weight matrix $p_{\theta_2}(\*W_2) = \mathcal{MN}(\*0, \*U^0_2, \*V^0_2)$. Now we can define the following generative model:
\begin{alignat*}{3}
&\*W_1 \sim \mathcal{MN}(\*0, \*U^0_1, \*V^0_1); \quad \*W_2 \sim \mathcal{MN}(\*0, \*U^0_2, \*V^0_2)\\
& \quad \*B = \*X\*W_1; \quad \*F = \psi(\*B)\*W_2\\
& \qquad \*Y \sim \mathcal{MN}(\*F, \tau^{-1}\*I_N, \*I_{D_y}) 
\end{alignat*}
where $\psi(\cdot)$ is a nonlinearity and $\mathcal{MN}(\*F, \tau^{-1}\*I_N, \*I_{D_y})$ corresponds to an independent multivariate Gaussian over each column of $\*Y$, i.e. $p(\*Y|\*X) = \prod_{i=1}^{D_y}\mathcal{N}(\*y_i|\*f_i, \tau^{-1}\*I_{N})$, where $\*f_i$ is a column of $\*F$\footnote{Note that this is just a simplifying assumption and not a limitation for our method. We could instead also model the correlations among the output variables $\*Y$ if we used a full covariance $\*C_{D_y}$ instead of $\*I_{D_y}$.}. Now if we make use of the matrix variate Gaussian property~\ref{eq:mnh} we have that the generative model becomes:
\begin{alignat*}{3}
\*B | \*X & \sim \mathcal{MN}(\*0, \*X\*U^0_1\*X^T, \*V^0_1)\\
\*F | \*B & \sim \mathcal{MN}(\*0, \psi(\*B)\*U^0_2\psi(\*B)^T, \*V^0_2)\\
\*Y | \*F & \sim \mathcal{MN}(\*F, \tau^{-1}\*I_N, \*I_{D_y})
\end{alignat*}
or else equivalently:
\begin{alignat*}{3}
\vect{(\*B)} | \*X & \sim \mathcal{N}(\*0, \hat{\*K}_{\theta_1}(\*X, \*X))\\
\vect{(\*F)} |\*B & \sim \mathcal{N}(\*0, \hat{\*K}_{\theta_2}(\*B, \*B))\\
\vect{(\*Y)} | \*F & \sim \mathcal{N}(\vect(\*F), \tau^{-1} (\*I_N\otimes\*I_{D_y}))
\end{alignat*}
where $\hat{\*K}_\theta(\*z_1, \*z_2) = \*K_{out}\otimes\*K_{in}(\*z_1,\*z_2;\*U) = \*V \otimes \big(\psi(\*z_1)\*U\psi(\*z_2)^T\big)$\footnote{$\psi(\cdot)$ is the identity function for the input layer.}. In other words, we have a composition of GPs where the covariance of each GP is governed by a kernel function of a specific form; it is the kroneker product of a global output and an input dependent kernel function, where the latter is composed of fixed dimension nonlinear basis functions (the inputs to each layer) weighted by their covariance. Essentially this kernel provides a distribution for each layer that is similar to a (correlated) multi-output GP, which was previously explored in the context of shallow GPs~\cite{yu2006stochastic,bonilla2007kernel,bonilla2007multi}. Therefore, in order to obtain the marginal likelihood of the targets $\*Y$ we have to marginalize over the function values $\*B$ and $\*F$, which results into a deep GP~\cite{damianou2012deep} with the aforementioned kernel function for each GP:
\begin{alignat*}{3}
&\log p(\*Y|\*X) = \\&\qquad \log\E_{p_{\theta_1}(\*B|\*X) p_{\theta_2}(\*F|\*B)}\big[\mathcal{N}(\vect(\*F), \tau^{-1}(\*I_N \otimes \*I_{D_y}))\big]
\end{alignat*}
A similar scenario was also considered theoretically in~\cite{duvenaud2014avoiding}. Now in order to obtain the posterior distribution of the parameters $\*W$ we will perform variational inference. We place a matrix variate Gaussian posterior distribution over the weights of the neural network, i.e. $q_{\phi_1}(\*W_1)q_{\phi_2}(\*W_2) = \mathcal{MN}(\*M_1, \*U_1, \*V_1)\mathcal{MN}(\*M_2, \*U_2, \*V_2)$, and the marginal likelihood lower bound in eq.~\ref{eq:marginal_lb} becomes:
\begin{alignat}{1}
&\mathcal{L}(\phi_{1,2}, \theta_{1,2})\leq \nonumber\\&\quad \E_{\tilde{p}(\*x,\*y)}\bigg[\E_{q_{\phi_1}(\*W_1)q_{\phi_2}(\*W_2)}\big[\log p(\*Y|\*X, \*W_1, \*W_2)\big] - \nonumber\\&\qquad-\sum_{i=1}^{2} KL(q_{\phi_i}(\*W_i)||p_{\theta_i}(\*W_i))\bigg]\nonumber\\
& \quad = \E_{\tilde{p}(\*x, \*y)}\bigg[L_{(\*X,\*Y)}(\phi_1, \phi_2) + \sum_{i=1}^{2} L_c(\phi_i,\theta_i)\bigg]
\end{alignat}
Noting that $\*Y$ only depends on $\*X,\*W_1,\*W_2$ through $\*F=\psi(\*B)\*W_2 = \psi(\*X\*W_1)\*W_2$, and applying the reparametrization trick, i.e.
\begin{alignat*}{1}
&\int q_{\phi_1}(\*W_1)q_{\phi_2}(\*W_2)\log p(\*Y|\*F(\*X, \*W_1, \*W_2))d\*W_{1,2} =\\ 
&\int \tilde{q}_{\phi_1}(\*B|\*X)\tilde{q}_{\phi_2}(\*F|\*B)\log p(\*Y|\*F)d\*B d\*F
\end{alignat*}
where (using \ref{eq:mnh}), 
\begin{align*}
\tilde{q}_{\phi_1}(\*B|\*X) & = \mathcal{N}(\vect(\!\mu_{\phi_1}(\*X)), \hat{\*K}_{\phi_1}(\*X, \*X))\\
\tilde{q}_{\phi_2}(\*F|\*B) & = \mathcal{N}(\vect(\!\mu_{\phi_2}(\*B)), \hat{\*K}_{\phi_2}(\*B, \*B))
\end{align*}
where $\phi_1 = (\*M_1, \*U_1, \*V_1)$, $\phi_2 = (\*M_2, \*U_2, \*V_2)$ are the variational parameters and $\!\mu_\phi(\*z) = \psi(\*z)\*M$ is the mean function. As we can see, $\tilde{q}_{\phi_1}(\*B|\*X), \tilde{q}_{\phi_2}(\*F|\*B)$ can be considered as approximate posterior GP functions while the local reparametrization trick provides the connection between the primal and dual GP view of the model. The variational objective thus becomes:
\begin{alignat}{1}
\mathcal{L}(\phi_{1,2}, \theta_{1,2}) & \leq \E_{\tilde{p}(\*x,\*y)}\bigg[\E_{\tilde{q}_{\phi_1}(\*B|\*X)\tilde{q}_{\phi_2}(\*F|\*B)}\big[\log p(\*Y|\*F)\big] + \nonumber\\&\qquad + \sum_{i=1}^{2}L_c(\phi_i, \theta_i)\bigg]
\label{eq:lb_dngp}
\end{alignat}

\subsection{Efficient sampling and pseudo-data}
Sampling distribution~\ref{eq:mnh} for every layer is however computationally intensive as we have to calculate the square root of the row covariance $\*K_{in}(\*A,\*A; \*U) = \*A\*U\*A^T$ (which has a cubic cost w.r.t. the amount of datapoints in $\*A$) every time. A simple solution is to only use its diagonal for sampling. This corresponds to samples from the marginal distribution of each pre-activation latent variable $\*b_i$ in the minibatch $\*A$. More specifically, we have that $\*b_i$ follows a multivariate Gaussian distribution where the covariance is controlled by two sources: the local scalar row variance (i.e. per datapoint feature correlations) and the global column, i.e. pre-activation latent variable (or target variable in the case of the output layer), covariance: $p(\*b_i|\*a_i) = \mathcal{N}(\*a_i\*M, \big(\*a_i\*U\*a_i^T\big) \odot \*V)$.

Despite its simplicity however this approach does not use the Gaussian Process nature of our model. In order to fully utilize this property we adopt an idea from the GP literature: the concept of pseudo-data~\cite{snelson2005sparse}. More specifically, we introduce pseudo inputs $\tilde{\*A}$ and pseudo outputs $\tilde{\*B}$ for each layer in the network and sample the distribution of each pre-activation latent variable $\*b_i$ conditioned on the pseudo-data:
\begin{alignat}{3}
p(\*b_i|\*a_i, \tilde{\*A}, \tilde{\*B})=& \mathcal{N}\bigg(\*a_i\*M + \!\sigma_{12}^T\!\Sigma_{11}^{-1}\big(\tilde{\*B} - \tilde{\*A}\*M\big), \nonumber\\&\qquad\big(\sigma_{22} - \!\sigma_{12}^T\!\Sigma_{11}^{-1}\!\sigma_{12}\big) \odot \*V\bigg)
\end{alignat}
where each of the covariance terms can be estimated as:
\begin{alignat}{3}
& \!\Sigma_{11} = \tilde{\*A}\*U\tilde{\*A}^T; \ & \!\sigma_{12} = \tilde{\*A}\*U\*a_i^T; \ & \sigma_{22} = \*a_i\*U\*a_i^T
\end{alignat}
As can be seen, the pseudo-data directly affect the distribution of each pre-activation latent variable: if the inputs are similar to the pseudo-inputs then the variance of the latent variable $\*b_i$ decreases and the mean is shifted towards the pseudo-data. This allows each layer in the network to be more certain in particular regions of the input space. However, if the inputs are not similar to the pseudo-inputs then the distribution of $\*b_i$ depends mostly on the parameters of the underlying matrix variate Gaussian posterior. 

It should be noted that the amount of pseudo-data for each layer $N_p$ should be $N_p < D$, where $D$ is the dimensionality of the input, as we are using a linear kernel for the row covariance (that becomes non-linear via the neural network nonlinearities) that has finite rank $D$. This enforces that the pseudo-data combined with a real input $\*a_i$ provide a positive definite kernel $\hat{\*K}$ for the joint Gaussian output distribution $p(\tilde{\*B}, \*b_i | \tilde{\*A}, \*a_i)$. Furthermore, we also "dampen" $\!\Sigma_{11}$ by adding to it a small diagonal matrix $\sigma^2 \*I$ where $\sigma^2 = 1e^{-8}$. This corresponds to assuming "noisy" pseudo-observations $\tilde{\*B}$~\cite{rasmussen2006gaussian} (where the noise is i.i.d.  from $\mathcal{N}(0, \sigma^2)$) which helps avoiding numerical instabilities during optimization (this is particularly helpful with limited precision floating-point).

At first glance it might seem that we now overparametrize each neural network layer, however  in practice this does not seem to be the case. From our experience relatively few pseudo-data per layer (compared to the input dimensionality) are necessary for increased performance. This still yields less parameters than fully factorized Gaussian posteriors. In addition, note that with the pseudo data formulation we could also assume that the weight posterior has zero mean $\*M = \*0$ (in GP parlance this corresponds to removing the mean function); this would reduce the number of parameters even further and still provide a useful model. This assumption leads to sampling the following distribution:
\begin{alignat}{3}
&p(\*b_i|\*a_i, \tilde{\*A}, \tilde{\*B}) =\nonumber\\ &\qquad \mathcal{N}\bigg(\!\sigma_{12}^T\!\Sigma_{11}^{-1}\tilde{\*B}, \big(\sigma_{22} - \!\sigma_{12}^T\!\Sigma_{11}^{-1}\!\sigma_{12}\big) \odot \*V\bigg)
\end{alignat}

Finally, since we want a fully Bayesian model, we also place fully factorized multiplicative Gaussian posteriors on both $\tilde{\*A}$ and $\tilde{\*B}$ along with log-uniform priors, as it was described in~\cite{kingma2015variational}. The final form of the bound~\ref{eq:lb_dngp} with the inclusion of the pseudo-data is:
\begin{alignat}{1}
&\mathcal{L}(\phi_{1,2}, \theta_{1,2}) \leq \nonumber \\&
\E_{\tilde{p}(\*x,\*y)}\bigg[\E_{q_{\phi_1}(\*B,\tilde{\*A}_1,\tilde{\*B}_1|\*X)q_{\phi_2}(\*F,\tilde{\*A}_2,\tilde{\*B}_2|\*B)}\big[\log p(\*Y|\*F)\big] + \nonumber\\&\qquad + \sum_{i=1}^{2}L_c(\phi_i, \theta_i)\bigg]
\label{eq:final_lb_dngp}
\end{alignat}
where:
\begin{alignat*}{1}
&q_{\phi_i}(\*B,\tilde{\*A},\tilde{\*B}|\*X) = \tilde{q}_{\phi_i}(\*B|\*X,\tilde{\*A},\tilde{\*B})q_{\phi_i}(\tilde{\*A})q_{\phi_i}(\tilde{\*B})\\
&L_c(\phi_i,\theta_i) = -KL(q(\*W_i)||p(\*W_i)) - \\&\qquad - KL(q(\tilde{\*A}_i)||p(\tilde{\*A}_i)) - KL(q(\tilde{\*B}_i)||p(\tilde{\*B}_i)) 
\end{alignat*}
where now $\phi_i$, $\theta_i$ also include the parameters of the distributions of the pseudo-data. The KL-divergence for these can be found at~\cite{kingma2015variational}. We can thus readily optimize the marginal likelihood lower bound of eq.~\ref{eq:final_lb_dngp} w.r.t. the parameters of the posterior and the pseudo data with stochastic gradient ascent. 

\subsection{Computational complexity}
A typical variational Bayesian neural network with a fully factorized Gaussian posterior sampled ``locally''~\cite{kingma2015variational} has asymptotic per-datapoint time complexity $\mathcal{O}(D^2)$ for the mean and variance in each layer, where $D$ is the input/output dimensionality. Our model adds the extra cost of inverting $\!\Sigma_{11}^{-1}$, that has cubic complexity with respect to the amount of pseudo-data $M$ for each layer. Therefore the asymptotic time complexity is $\mathcal{O}(D^2 + M^3)$ and since usually $M << D$, this does not incur a significantly extra computational cost.

%% file: related.tex
\cite{graves2011practical} firstly introduced a practical way of variational inference for neural networks. Despite the fact that the proposed (biased) estimator had good performance on a recurrent neural network task, it was not as effective on the regression task of~\cite{hernandez2015probabilistic}.~\cite{blundell2015weight} proposed to use an alternative unbiased estimator that samples on the relatively high variance weight space but nonetheless provided good performance on a reinforcement learning task.~\cite{kingma2015variational} subsequently presented the ``local reparametrization trick", which makes use of Gaussian properties so as to sample in the function space, i.e. the hidden units. This provides both reduced memory requirements as well as reduced variance for the expected log-likelihood estimator. However, for their model they still use a fully factorized posterior distribution that doubles the amount of parameters in each layer and does not allow the incorporation of pseudo-data.

\cite{gal2015dropout} also provides connections between Bayesian neural networks and deep Gaussian processes, but they only consider independent Gaussians for each column of the weight matrix (which in our case correspond to $p(\*W)=\mathcal{MN}(\*M, \!\sigma^2\*I, \*I)$) and do not model the variances of the hidden units. Furthermore the approximating variational distribution is quite limited as it corresponds to simple Bernoulli noise and delta approximating distributions for the weight matrix: it is a mixture of two delta functions for each column of the weight matrix, one at zero and the other at the mean of the Gaussian. This is in contrast to our model where we can explicitly learn the (possibly non-diagonal) covariance for both the input and output dimensions of each layer through the matrix variate Gaussian posterior. In addition, sampling is done in the weight space and not the function space as in our model, thus preventing the use of pseudo-data. 

Finally, \cite{hernandez2015probabilistic} also assume fully factorized posterior distributions and uses Expectation Propagation~\cite{minka2001expectation} instead of variational inference. Closed form approximations bypass the need for sampling in the model, which in turn makes it easier to converge. However their derivation is limited to rectified linear nonlinearities and regression problems, thus limiting the applicability of their model. Furthermore, since each datapoint is treated as new during the update of the parameters, special care has to be given so as to not perform a lot of passes through the dataset since this will in general shrink the variances of the weights of the network.

%% file: experiments.tex
All of the models were coded in Theano~\citep{bergstra2010theano} and optimization was done with Adam~\cite{kingma2014adam}, using the default hyper-parameters and temporal averaging. We parametrized the prior for each weight matrix as $p(\*W) = \mathcal{MN}(\*0, \*I, \*I)$ unless stated otherwise. Following~\cite{hernandez2015probabilistic} we also divide the input to each layer (both real and pseudo) by the square root of its dimensionality so as to keep the scale of the output (before the nonlinearity) independent of the incoming connections. We used rectified linear units~\cite{nair2010rectified} (ReLU) and we initialized the mean of each matrix variate Gaussian via the scheme proposed in~\cite{he2015delving}. For the initialization of the pseudo-data we sampled the entries of $\tilde{\*A},\tilde{\*B}$ from $\mathcal{U}[-0.01, 0.01]$. 
We used one posterior sample to estimate the expected log-likelihood before we update the parameters.

We test under two different scenarios: regression and classification. For the regression task we experimented with the UCI~\cite{asuncion2007uci} datasets that were used in ``Probabilistic Backpropagation" (PBP)~\cite{hernandez2015probabilistic} and in ``Dropout as a Bayesian Approximation"~\cite{gal2015dropout}. For the classification task we evaluated our model on the permutation invariant MNIST benchmark dataset, so as to compare against other popular neural network models. 

Finally we also performed a toy regression experiment on the same artificially generated data as~\cite{hernandez2015probabilistic}, so that we can similarly visualize the predictive distribution that our model provides.

\subsection{Regression experiments}
For the regression experiments we followed a similar experimental protocol with~\cite{hernandez2015probabilistic}: we randomly keep 90\% of the dataset for training and use the remaining to test the performance. This process is repeated 20 times (except from the ``Protein" dataset where it is performed 5 times and the ``Year" dataset where it is performed once) and the average values along with their standard errors are reported at Table~\ref{tab:reg_res}. Following~\cite{hernandez2015probabilistic} we also introduce a Gamma prior, $p(\tau) = Gam(a_0=6,b_0=6)$ and posterior $q(\tau) = Gam(a_1, b_1)$ for the precision of the Gaussian likelihood and we parametrized the matrix variate Gaussian prior for each layer as $p(\*W) = \mathcal{MN}(\*0, \tau_r^{-1}\*I, \tau_c^{-1}\*I)$, where $p(\tau_r), p(\tau_c) = Gam(a_0=1,b_0=0.5)$ and $q(\tau_r)q(\tau_c) = Gam(a_r, b_r)Gam(a_c, b_c)$\footnote{For this choice of distribution both $\E_{q(\tau_r)q(\tau_c)}[KL(q(\*W|\*M, \*U, \*V)||p(\*W|\*0, \tau_r^{-1}\*I, \tau_c^{-1}\*I))]$ and the KL-divergence between $q(\tau_r)q(\tau_c)$ and $p(\tau_r)p(\tau_c)$ can be computed in closed form.}. We optimized $a_1,b_1,a_r,b_r,a_c,b_c$ along with the remaining variational parameters. We do not use a validation set and instead train the networks up until convergence in the training set. We use one hidden layer of 50 units for all of the datasets, except for the larger ``Protein'' and ``Year'' datasets where we use 100 units. We normalized the inputs $\*x$ of the network to zero mean and unit variance but we did not normalize the targets $\*y$. Instead we parametrized the network output as $\*y = f(\*x)\odot\!\sigma_y + \!\mu_y$ where $f(\cdot)$ represents the neural network and $\!\mu_y, \!\sigma_y$ are the, per-dimension, mean and standard deviation of the target variable, estimated from the training set. Similarly to~\cite{gal2015dropout} we set the upper bound of the variational dropout rate to $0.005$, $0.05$ and we used 10 pseudo-data pairs for each layer for all of the datasets, except for the smaller ``Yacht" dataset where we used 5 and the bigger ``Protein" and ``Year" where we used 20.

\begin{table*}[htb]
\centering
\resizebox{2.07\columnwidth}{!}{%
    \begin{tabular}{lccccccccc}
    & \multicolumn{4}{c}{Avg. Test RMSE and Std. Errors} & \multicolumn{4}{c}{Avg. Test LL and Std. Errors}\\
        \textbf{Dataset} &  \multicolumn{1}{c}{\textbf{VI}} & \multicolumn{1}{c}{\textbf{PBP}} & \multicolumn{1}{c}{\textbf{Dropout}} & \multicolumn{1}{c}{\textbf{VMG}} & \multicolumn{1}{c}{\textbf{VI}} & \multicolumn{1}{c}{\textbf{PBP}} & \multicolumn{1}{c}{\textbf{Dropout}} & \multicolumn{1}{c}{\textbf{VMG}} \\\hline
        Boston & 4.32$\pm$0.29 & 3.01$\pm$0.18 & 2.97$\pm$0.85 &\textbf{2.70$\pm$0.13} & -2.90$\pm$0.07 & -2.57$\pm$ 0.09 & -2.46$\pm$0.25 & \textbf{-2.46$\pm$0.09}\\
        Concrete &7.19$\pm$0.12& 5.67$\pm$0.09 & 5.23$\pm$ 0.53 & \textbf{4.89$\pm$0.12} & -3.39$\pm$0.02 &-3.16$\pm$0.02 & -3.04$\pm$0.09&\textbf{-3.01$\pm$0.03}\\
        Energy & 2.65$\pm$0.08& 1.80$\pm$0.05 & 1.66$\pm$0.19 & \textbf{0.54$\pm$0.02} & -2.39$\pm$0.03& -2.04$\pm$0.02 & -1.99$\pm$0.09 & \textbf{-1.06$\pm$0.03}\\
        Kin8nm  & 0.10$\pm$0.00 & 0.10$\pm$0.00& 0.10$\pm$0.00&\textbf{0.08$\pm$0.00} & 0.90$\pm$0.01 & 0.90$\pm$0.01 & 0.95$\pm$0.03 &\textbf{1.10$\pm$0.01}\\
        Naval & 0.01$\pm$0.00 & 0.01$\pm$0.00 & 0.01$\pm$0.00 &\textbf{0.00$\pm$0.00} &3.73$\pm$0.12& 3.73$\pm$0.01 & \textbf{3.80$\pm$0.05}& 2.46$\pm$0.00 \\
        Pow. Plant  &4.33$\pm$0.04 & 4.12$\pm$0.03 & \textbf{4.02$\pm$0.18} & 4.04$\pm$0.04 & -2.89$\pm$0.01 & -2.84$\pm$0.01 & \textbf{-2.80$\pm$0.05} &-2.82$\pm$0.01\\
        Protein &4.84$\pm$0.03 & 4.73$\pm$0.01 & 4.36$\pm$0.04 & \textbf{4.13$\pm$0.02}&-2.99$\pm$0.01&-2.97$\pm$0.00&-2.89$\pm$0.01 &\textbf{-2.84$\pm$0.00}\\ 
        Wine & 0.65$\pm$0.01 & 0.64$\pm$0.01 & \textbf{0.62$\pm$0.04} & 0.63$\pm$0.01 &-0.98$\pm$0.01 & -0.97$\pm$0.01 & \textbf{-0.93$\pm$0.06} & -0.95$\pm$0.01\\
        Yacht &6.89$\pm$0.67& 1.02$\pm$0.05&1.11$\pm$0.38&\textbf{0.71$\pm$0.05}&-3.43$\pm$0.16 &-1.63$\pm$0.02& -1.55$\pm$0.12&\textbf{-1.30$\pm$0.02}\\
        Year &9.034$\pm$NA& 8.879$\pm$NA & 8.849$\pm$NA & \textbf{8.780$\pm$NA} &-3.622$\pm$NA & -3.603$\pm$NA&\textbf{-3.588$\pm$NA} &-3.589$\pm$NA \\\hline
    \end{tabular}%
   }
    \caption{Average test set RMSE, predictive log-likelihood and standard errors for the regression datasets. VI, PBP and Dropout correspond to the variational inference method of~\cite{graves2011practical}, probabilistic backpropagation~\cite{hernandez2015probabilistic} and dropout uncertainty~\cite{gal2015dropout}. VMG (Variational Matrix Gaussian) corresponds to the proposed model.}
    \label{tab:reg_res}
\end{table*}

As we can see from the results at Table~\ref{tab:reg_res} our model overall provides lower root mean square errors, compared to VI~\cite{graves2011practical}, PBP~\cite{hernandez2015probabilistic} and Dropout~\cite{gal2015dropout} on most datasets. In addition, we also observe better performance according to the predictive log-likelihoods; our model outperforms VI and PBP on most datasets and is better than Dropout on 6 out of 10. These results empirically verify the effectiveness of our model: with the matrix variate Gaussian posteriors along with the Gaussian Process interpretation we have a model that is flexible and consequently can both better fit the data, and, in the case of the predictive log-likelihoods, make an accurate estimation of the predictive uncertainty. 

\subsection{Classification experiments}
For the classification experiments we trained networks with a varying number of layers and hidden units per layer. We used the last 10000 samples of the training set as a validation set for model selection, minibatches of 100 datapoints and set the upper bound for the variational dropout rate to $0.25$. We used the same amount of pseudo-data pairs for each layer, but tuned those according to the validation set performance (we set an upper bound of 150 pseudo-data pairs per layer). We did not use any kind of data augmentation or preprocessing. The results can be seen at Table~\ref{tab:mnist_res}.
\begin{table}[h]
\centering
\resizebox{\columnwidth}{!}{%
\begin{tabular}{l|l|c}
\textbf{Method} & \textbf{\# layers} & \textbf{Test err.}\\\hline
Max. Likel.~\cite{simard2003best} & 2$\times$800 & 1.60\\
Dropout~\cite{srivastava2013improving} & - & 1.25 \\
DropConnect~\cite{wan2013regularization} & 2$\times$800 & 1.20 \\\hline
Bayes B. SM~\cite{blundell2015weight} & 2$\times$400 & 1.36\\
 &2$\times$800 & 1.34\\
 &2$\times$1200 & 1.32\\\hline
Var. Dropout~\cite{kingma2015variational} & 3$\times$150 & $\approx$ 1.42\\
&3$\times$250 & $\approx$ 1.28\\
&3$\times$500 & $\approx$ 1.18\\
&3$\times$750 & $\approx$ 1.09\\\hline
VMG & 2$\times$400 & \textbf{1.15} \\
 & 3$\times$150 & 1.18\\
 & 3$\times$250 & 1.11\\
 & 3$\times$500 & 1.08\\
 & 3$\times$750 & \textbf{1.05}\\\hline
\end{tabular}%
}
\caption{Test errors for the permutation invariant MNIST dataset. Bayes B. SM correspond to Bayes by Backprop with the scale mixture prior and the variational dropout results are from the Variational (A) model that doesn't downscale the KL-divergence (so as to keep the comparison fair).
}
\label{tab:mnist_res}
\end{table}

As we can observe our Bayesian neural network performs better than other popular neural networks models for small network sizes. For example, with only three hidden layers of 150 units it achieves 1.18\% test error on MNIST, a result that is better than maximum likelihood~\cite{simard2003best}, Dropout~\cite{srivastava2013improving}, DropConnect~\cite{wan2013regularization} and Bayes by Backprop~\cite{blundell2015weight}, where all of the aforementioned methods have significantly bigger architectures than our model. Furthermore, it is also significantly better than a neural network of the same size trained with variational dropout~\cite{kingma2015variational}. We can probably attribute this effect to the Gaussian Process property; for regular neural networks a small network size means that there are not enough parameters to learn an effective classifier. This is in contrast to our model where through the learned pseudo-data we can maintain this property and consequently increase the flexibility of the model and compensate for the lack of network size.

\subsection{Toy experiment}
\begin{figure}[ht]
\centering
    \subfigure[Factorized Gaussian]{\includegraphics[height=1.in]{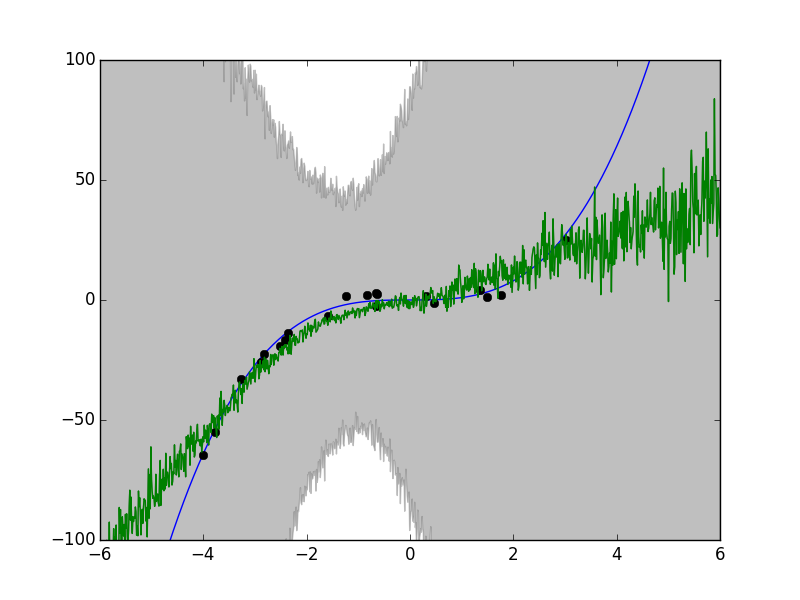}}
    ~
    \subfigure[PBP]{\includegraphics[height=1.in]{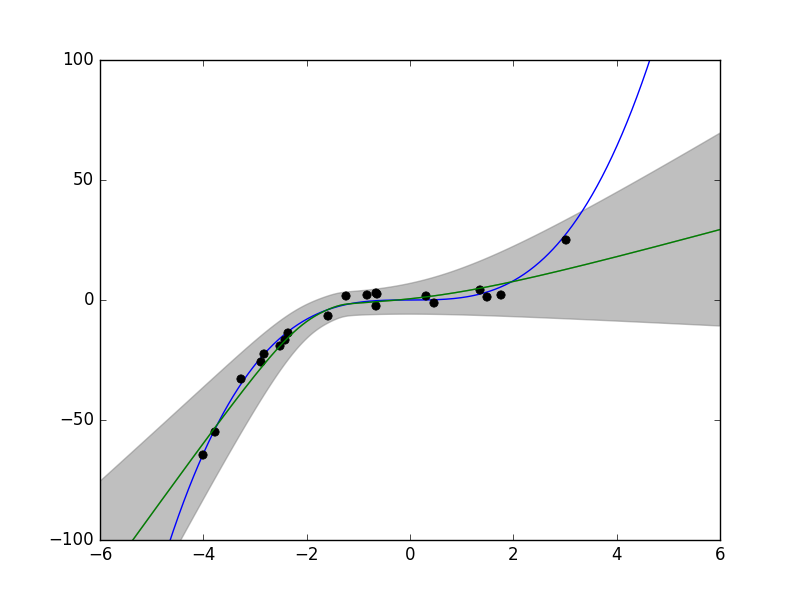}}\\
    \subfigure[Dropout]{\includegraphics[height=1.in]{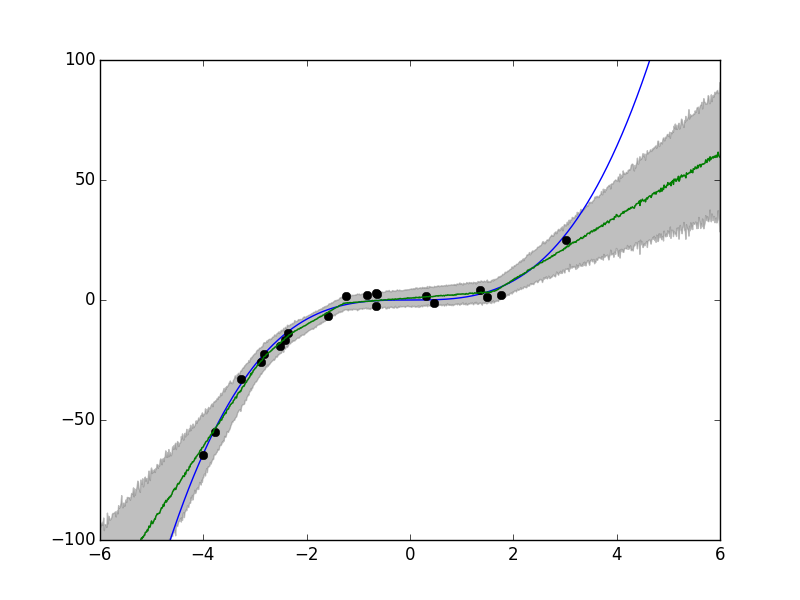}}
    ~
    \subfigure[Matrix Gaussian]{\includegraphics[height=1.in]{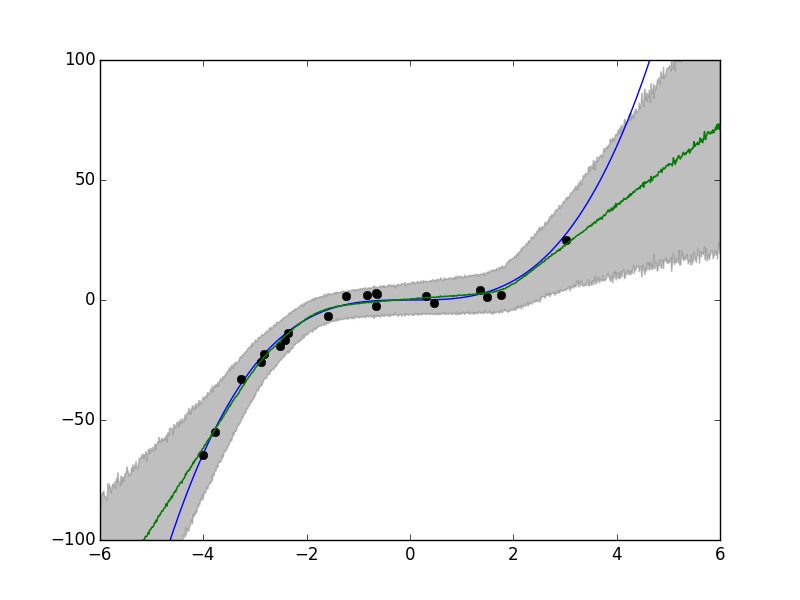}}%
    \caption{Predictive distributions for the toy dataset. Grey areas correspond to $\pm 3$ standard deviations around the mean function.}
    \label{fig:toy_pred}
\end{figure}

In order to visually access the quality of the uncertainty that our model provides, we also performed an experiment on the simple toy dataset that was used in~\cite{hernandez2015probabilistic}. We sampled 20 inputs $x$ from $\mathcal{U}[-4, 4]$ and parametrized the target variable as $y_n = x_n^3 + \epsilon_n$ where $\epsilon_n \sim N(0, 9)$. We then fitted a neural network with matrix Gaussian posteriors (with diagonal covariance matrices), a neural network that had a fully factorized Gaussian distribution for the weights and a dropout network. All of the networks had a single hidden layer of 100 units. For our model we used two pseudo-data pairs for the input layer, four for the output layer and set the upper bound of the variational dropout rate to $0.2$. The dropout rate for the dropout network was zero for the input layer and $0.2$ for the hidden layer. The resulting predictive distributions (after 200 samples) can be seen in figure~\ref{fig:toy_pred} (with three standard deviations around the mean).

As we can see the network with matrix Gaussian posteriors provides a realistic predictive distribution that seems slightly better compared to the one obtained from PBP~\cite{hernandez2015probabilistic}. Interestingly, the simple fully factorized Gaussian (sampled with the ``local reparametrization trick") neural network failed to obtain a good fit for the data as it was severely underfitting due to the limited amount of datapoints. This resulted into a very uncertain and noisy predictive distribution that vaguely captured the mean function. This effect is not observed with our model; with the Gaussian Process property we effectively increase the flexibility of our model thus allowing the weight posterior to be closer to the prior without severe loss in performance. Furthermore, we only have a handful of variance parameters to learn, which consequently provides easier and more robust posterior uncertainty estimation. Finally, it seems that the dropout network provides a predictive distribution that is slightly ``overfitted'' as the confidence intervals do not diverge as heavily in areas where there are no data.

%% file: conclusion.tex
We introduce a scalable variational Bayesian neural network where the parameters are governed by a probability distribution over random matrices: the matrix variate Gaussian. By utilizing properties of this distribution we can see that our model can be considered as a composition of Gaussian Processes with nonlinear kernels of a specific form. This kernel is formed from the kroneker product of two separate kernels; a global output kernel and an input specific kernel, where the latter is composed from fixed dimension nonlinear basis functions (the inputs to each layer) weighted by their covariance. We continue in exploiting this duality and introduce pseudo input-output pairs for each layer in the network, which in turn better maintain the Gaussian Process properties of our model thus increasing the flexibility of the posterior distribution.

We tested our model in two scenarios: the same regression task as PBP~\cite{hernandez2015probabilistic} and Dropout uncertainty~\cite{gal2015dropout} and the benchmark permutation invariant MNIST classification task. For the regression task we found that our model overall achieves better RMSE and predictive log-likelihoods than VI~\cite{graves2011practical}, PBP and Dropout uncertainty. For the classification task we found that our model provides better errors than state of the art methods for small architectures. This demonstrates the effectiveness of the Gaussian Process property; with the pseudo-data we increase the flexibility of our model thus countering the fact that we have a limited capacity neural network. 

Finally, we also empirically verified the quality of the predictive distribution that our model provides on the same toy experiment as PBP~\cite{hernandez2015probabilistic}.

%% file: appendix.tex
\section{KL divergence between matrix variate Gaussian prior and posterior}
Let $\mathcal{MN}_0(\*M_0, \*U_0, \*V_0)$ and $\mathcal{MN}_1(\*M_1, \*U_1, \*V_1)$ be two matrix variate Gaussian distributions for random matrices of size $n\times p$. We can use the fact that the matrix variate Gaussian is a multivariate Gaussian if we flatten the matrix, i.e. $\mathcal{MN}_0(\*M_0, \*U_0, \*V_0) = \mathcal{N}_0(\vect(\*M_0), \*V_0 \otimes \*U_0)$, and as a result use the KL-divergence between two multivariate Gaussians:
\begin{align*}
    KL(\mathcal{N}_0 || \mathcal{N}_1) &= \frac{1}{2}\bigg(\tr(\!\Sigma_1^{-1}\!\Sigma_0) + \\& +(\!\mu_1 - \!\mu_0)^T\!\Sigma_1^{-1}(\!\mu_1 - \!\mu_0) - \\& -K + \log\frac{|\!\Sigma_1|}{|\!\Sigma_0|}\bigg) \\
     & =\frac{1}{2}\bigg(\tr\big((\*V_1\otimes\*U_1)^{-1}(\*V_0\otimes\*U_0)\big) + \\ & + \big(\vect(\*M_1) - \vect(\*M_0)\big)^T\\&\big(\*V_1\otimes\*U_1\big)^{-1}\big(\vect(\*M_1) - \vect(\*M_0)\big) -\\&- np + \log\frac{|\*V_1\otimes\*U_1|}{|\*V_0\otimes\*U_0|}\bigg)
\end{align*}
Now to compute each term in the KL efficiently we need to use some properties of the vectorization and Kronecker product:
\begin{align}
    t_a & = \tr\big((\*V_1\otimes\*U_1)^{-1}(\*V_0\otimes\*U_0)\big)\nonumber\\
     &= \tr\big((\*V_1^{-1}\otimes\*U_1^{-1})(\*V_0\otimes\*U_0)\big)\nonumber\\
    & = \tr\big((\*V_1^{-1}\*V_0)\otimes(\*U_1^{-1}\*U_0)\big)\nonumber\\
    & = \tr(\*U_1^{-1}\*U_0)\tr(\*V_1^{-1}\*V_0)
\end{align}
\begin{align}
    t_b & = \big(\vect(\*M_1) - \vect(\*M_0)\big)^T\big(\*V_1\otimes\*U_1\big)^{-1}\nonumber\\&\big(\vect(\*M_1) - \vect(\*M_0)\big) \nonumber\\ & = \vect(\*M_1 - \*M_0)^T(\*V_1^{-1}\otimes\*U_1^{-1})\vect(\*M_1 - \*M_0)\nonumber\\
    & = \vect(\*M_1 - \*M_0)^T \vect(\*U_1^{-1}(\*M_1 - \*M_0)\*V_1^{-1})\nonumber\\
    & = \tr\big((\*M_1 - \*M_0)^T\*U_1^{-1}(\*M_1 - \*M_0)\*V_1^{-1}\big)
\end{align}
\begin{align}
    t_c & = \log\frac{|\*V_1\otimes\*U_1|}{|\*V_0\otimes\*U_0|} \nonumber\\
    & = \log\frac{|\*U_1|^p |\*V_1|^n}{|\*U_0|^p |\*V_0|^n}\nonumber\\
    & = p\log|\*U_1| + n\log|\*V_1| -\nonumber\\&- p\log|\*U_0| - n\log|\*V_0|
\end{align}
So putting everything together we have that:
\begin{align}
    KL(\mathcal{MN}_0, \mathcal{MN}_1) & = \frac{1}{2}\bigg(\tr(\*U_1^{-1}\*U_0)\tr(\*V_1^{-1}\*V_0) + \nonumber\\& + \tr\big((\*M_1 - \*M_0)^T\*U_1^{-1}(\*M_1 - \*M_0)\*V_1^{-1}\big) - \nonumber\\ & - np + p\log|\*U_1| + n\log|\*V_1| - \nonumber\\&-p\log|\*U_0| - n\log|\*V_0|\bigg)    
\end{align}

\section{Different toy dataset}
\begin{figure}[ht]
\centering
    \subfigure[PBP]{\includegraphics[height=1.in]{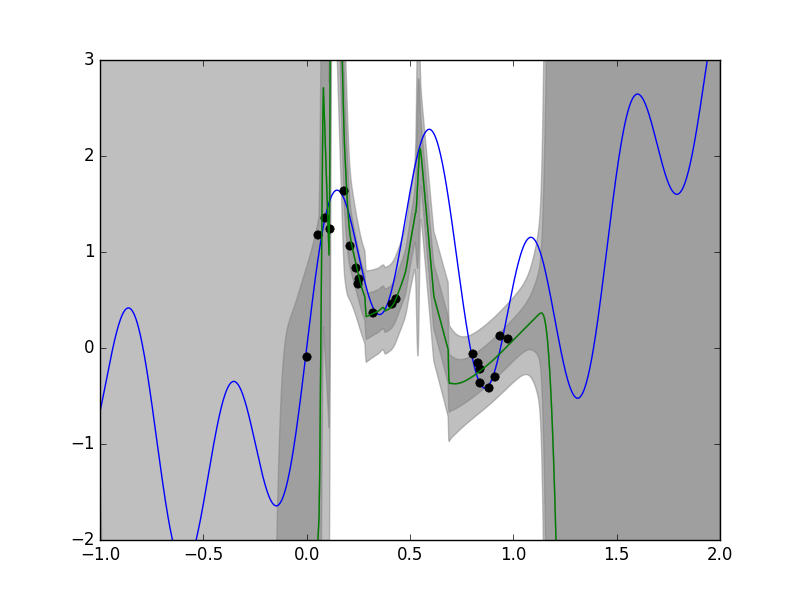}}
    ~
    \subfigure[MG ReLU]{\includegraphics[height=1.in]{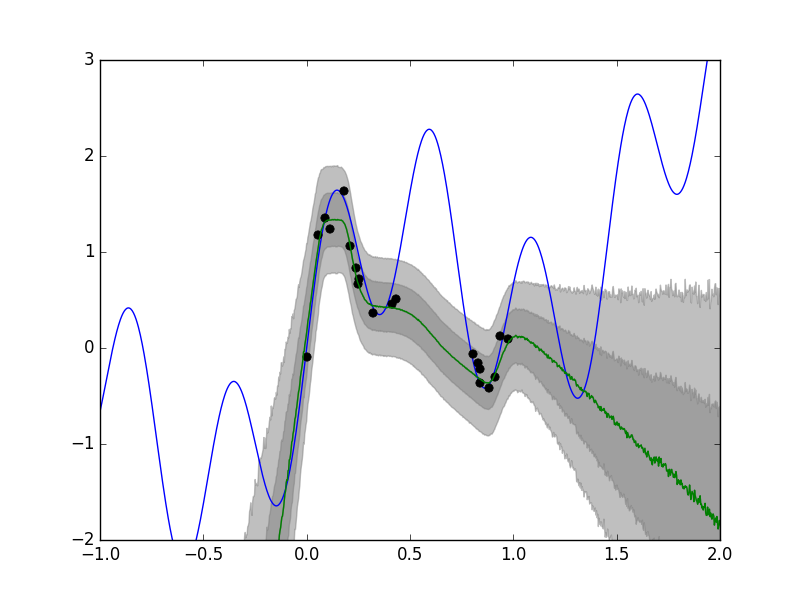}}\\
    \subfigure[MG cosine]{\includegraphics[height=1.in]{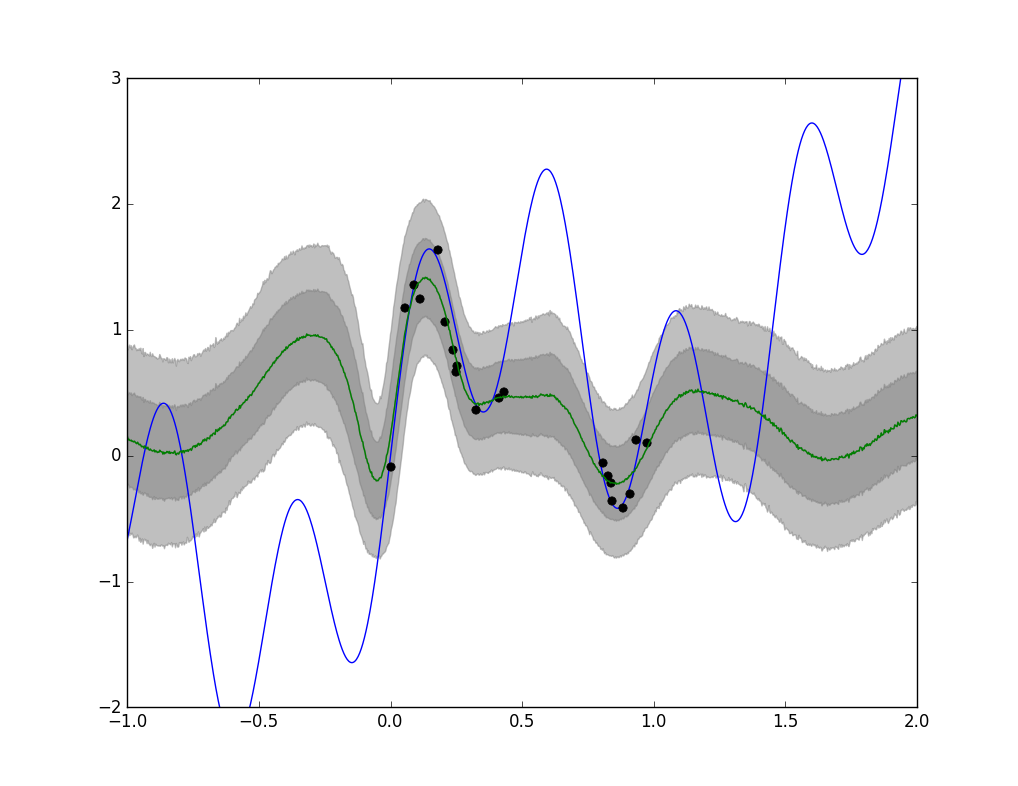}}
    ~
    \subfigure[MG tanh]{\includegraphics[height=1.in]{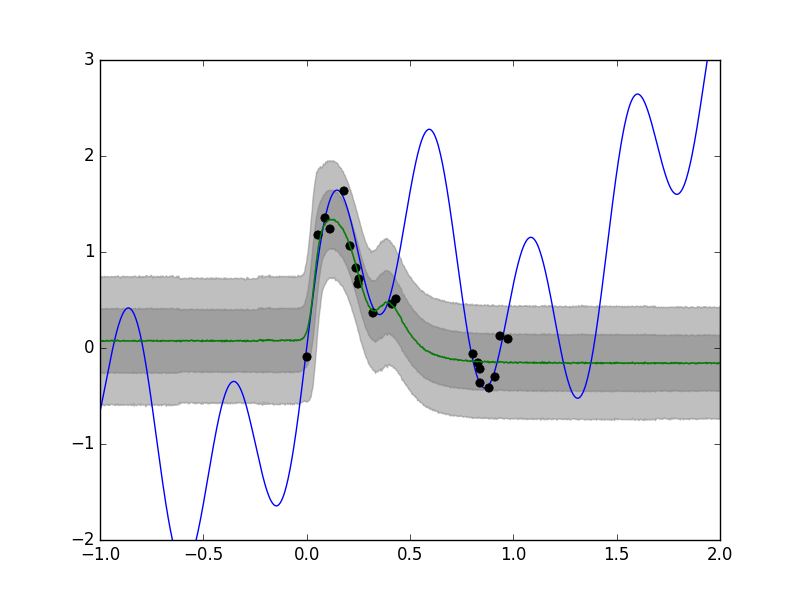}}%
    \caption{Predictive distributions for the toy dataset. Grey areas correspond to $\pm \{1, 2\}$ standard deviations around the mean function.}
    \label{fig:toy_pred_2}
\end{figure}

We also performed an experiment with a different toy dataset that was employed in~\cite{osband2016deep}. We generated 12 inputs from $U[0, 0.6]$ and 8 inputs from $U[0.8, 1]$. We then transform those inputs via:
\begin{align}
y_i = x_i + \epsilon_i + \sin(4 (x_i + \epsilon_i)) + \sin(13(x_i + \epsilon_i))\nonumber
\end{align}
where $\epsilon_i \sim \mathcal{N}(0, 0.0009)$. We continued in fitting four neural networks that had two hidden-layers with 50 units each. The first was trained with probabilistic back-propagation~\cite{hernandez2015probabilistic}, and the remaining three with our model while varying the nonlinearities among the layers: we used ReLU, cosine and hyperbolic tangent activations. For our model we set the upper bound of the variational dropout rate to $0.2$ and we used $2$ pseudo data pairs for the input layer and $4$ for the rest. The resulting predictive distributions can be seen at Figure~\ref{fig:toy_pred_2}.

